\documentclass[a4paper, conference]{ieeeconf}
\pdfoutput=1
\IEEEoverridecommandlockouts
\usepackage[bookmarks=false]{hyperref}
\usepackage{graphicx} 
\usepackage{subfigure}
\usepackage{xcolor}
\usepackage{multirow}
\usepackage{siunitx}
\usepackage{cleveref}
\usepackage{fullwidth}
\usepackage{comment}

\title{Detecting Parking Spaces in a Parcel using Satellite Images}

\author{Murugesan Vadivel, SelvaKumar Murugan, Suriyadeepan Ramamoorthy \\ Vaidheeswaran Archana, Malaikannan Sankarasubbu \hspace{1em}\\
Saama Technologies AI Research Lab \\
Chennai, India\\
{\tt\small \{murugesan.vadivel,suriyadeepan.ramamoorthy, selvakumar.murugan, malaikannan.sankarasubbu\}@saama.com}
}

\begin{document}

\maketitle
\thispagestyle{empty}
\pagestyle{empty}

\begin{abstract}
  Remote Sensing Images from satellites have been used in various domains for detecting and understanding structures on the ground surface. In this work, satellite images were used for localizing parking spaces and vehicles in parking lots for a given parcel using an RCNN based Neural Network Architectures.Parcel shape files and raster images from USGS image archive was used for developing images for both training and testing. Feature Pyramid based Mask RCNN yields average class accuracy of 97.56\% for both parking spaces and vehicles..
\end{abstract}

\section{Introduction}
Identifying and segmenting parking spaces is a well-studied problem in the computer vision community, with plenty of real-world applications. More often than not, the objective is to estimate the occupancy of parking spaces. Estimating the occupancy of parking spaces would allow the parking management to efficiently route traffic to, and from the premise, either autonomously or semi-autonomously and optimize parking spaces. The average occupancy estimate of parking spaces in a building, can be used to deduce the population density of the building.

The existing parking lot identification systems are highly accurate real-time detection systems which solve the problem of real-time occupancy estimation. Naturally these systems are quite hardware-specific, laborious, expensive to install and maintain. 

The existing systems can be broadly classified into two categories : Sensor-based systems and Software-based systems.

The Sensor-based systems are heavily dependent on locally placed sensors like Ultrasonic sensors \cite{ultrasensor} or laser sensor \cite{lasersensor} and they are supported by expensive hardware. 

Apart from sensor-based systems, end-to-end software-based systems also exist. For example, SIFT-based object detection is used for identifying parking spaces. 

Other systems use Convolutional Neural Networks \cite{cnn} for labeling individual spaces and returning number of empty spaces in an entire image.\cite{cite2}

Most of these systems do not take weather conditions into account. Others like Parking Space Classification need to train the model for different weather conditions. In addition to that, the presence of night stamp, a time stamp that indicates whether its day time or night time. Such a time stamp which could be used as a feature for classification, is also required.

The problem we are focusing on in this work is not a real time detection of parking spaces but for particular set of time series. In this work, we are proposing a methodology for detecting outdoor parking spaces in any given building on a given day, using satellite images.

We propose a system which relies on Satellite images, to identify parking spaces and estimate the occupancy. After experimenting with a series of experiment, we find that by using Mask Region based Convolutional Neural Networks (Mask-RCNN) \cite{mrcnn} enhanced by Feature Pyramid Networks (FPN) \cite{fpn} for identify parking spaces and vehicle is the satellite images is far more accurate than other architectures.

\section{Data}
Remote Sensing Images are totally different from the ordinary images that is been used in common computer vision problems. Remote Sensing sensors that are present in the satellites records the interactions of the electromagnetic radiation with an object that the energy strikes. The part of the Electromagnetic spectrum that we experience daily is visible light, however most of the EM spectrum falls outside the range of the relatively narrow portion that we can see with our eyes. Remote Sensing Imagery can be recorded in two different ways: one is Photographic/ analog in which remote sensing uses film to record reflected electromagnetic radiation to produce the image and another is the Digital in which the sensor records the reflected electromagnetic radiation that impacts the sensor in numerical values that can be interpreted as images \cite{satsensor}.

Two different types of data are necessary for the proposed solution:
\begin{itemize}
    \item \textbf{Rasters}: Rasters are digital aerial photographs, imagery from satellites, digital pictures, or even scanned maps. A raster dataset is composed of rows and columns of pixels known as cells. Each pixel represents a geographical region, and the value in that pixel represents some characteristics of that region. Raster data are mostly used in a GIS application when we want to display information that is continuous across an area and cannot easily be divided into vector features. The process of capturing raster data from an aeroplane or satellite is called remote sensing.
    \item \textbf{Shapefiles}: A shapefile is a simple, nontopological format for storing the geometric location and attribute information of geographic features. Geographic features in a shapefile can be represented by points, lines, or polygons (areas) and each feature has a set of associated attributes. For example, we are using parcel shapefile which has multiple polygon features for each parcel in a rasters and each polygon feature has some attributes like address of the parcel, total area, owned by and so on.
\end{itemize}

\subsection{Data Acquisition}

There are two main ways to collect Rasters images for any particular area:

\begin{itemize}
    \item \textbf{Image Archives from Open Access Platforms}: Several platforms like USGS Earth Explorer and ESA Sentinel Online collect data from multiple providers open source it. The advantages of using this method is that the rasters obtained are preprocessed by the removal of cloud coverage. Furthermore, it is easy to acquire information, given a coordinate, form these platforms. Finally, it supports the orthorectification \cite{orthorec}of images, which is the process of using cloud elevation data to correct the displacements caused by the difference in terrain and camera tilt. Orthorectification \cite{orthorec} is crucial in studying surface features. However, one of the main disadvantages of open access platforms is that we cannot access the latest image data from these platforms.
    \item \textbf{APIs}: Rasters can also be accessed through API’s of different image providers like Planet, Digital globe, and Skywatch. These platforms can be used to acquire real-time inference raster images, unlike the open access platforms. Furthermore, this helps to access images that are as recent as a couple of days or weeks with specified resolution and other characteristics like spectral resolution,..etc. 
    
\end{itemize}

We collected rasters using the first method through USGS Earth Explorer platform since we don't need real time images for training. Rasters only to some particular areas of California and Arizona have been collected based on the building areas.

\subsection{Data Preprocessing}
There are several challenges in preprocessing satellite images like the high resolution of images, mapped geospatial information and rotational invariance for raw raster images. This is why the preprocessing techniques employed for common image datasets like VOC \cite{voc}, COCO \cite{coco}cannot be used for Satellite Imagery. 

In the process of extracting satellite images in parcel level from the raster images, parcel shapefiles plays a vital role. The parcel shape files are used to separate parcel features using polygonal boundaries in the raster images. Those parcel shapefiles are collected from the official goverment site of california and pheonix state as we have the rasters only for those regions. All the preprocessing steps are detailed in the below steps.

\begin{itemize}
    \item Stitch all the rasters collected based on the states using GDAL \cite{gdal} toolkit.
    \item Visualize both the Stitched raster and the parcel shapefile using QGis \cite{qgis}(also other GIS data view like ArcGIS and others).
    \item Select the Parcel feature from the shapefile that has a parking lot in it using QGis and convert those selected features into a separate shapefile.
    \item Using the Fiona and Rasterio framework to load both the stitched raster and selected parcel shapefiles. Crop each and every selected parcel from the raster and save only the RBG bands as a PNG image.
\end{itemize}

\begin{figure}[h] 
\centering
\includegraphics[width=90mm, height=110mm]{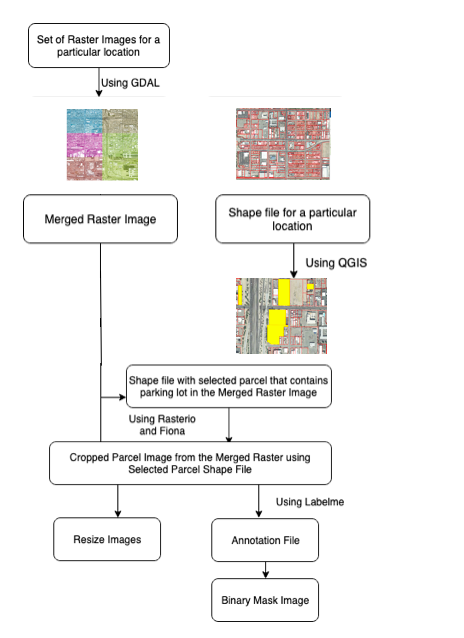}
\caption{Steps in Data Preprocessing}
\label{fig:model}
\end{figure}

\subsection{Dataset Annotation}
Annotation refers to the task of finding and labelling objects in images. In this work, images were annotated using the LabelMe \cite{labelme} tool. There are two different types of annotation involved in annotating images for object localization, one is using Bounding Box (Rectangle shape) and another is using Polygon (random shape). Annotation methods differs depends on the type of object localization whether it is detection using bbox or pixel-wise segmentation. As we are about to try both detection using bounding box and segmentation, polygon based annotation would serve better. Bounding box can be obtained from the annotated polygon by extracting the maximum (x,y) from the annotated polygon. 
\subsection{Data Augmentation}
One of the main drawbackS of adding more constraints such as location and parcel with parking lots for selecting the raster and the parcel feature is that the training parcel image will be sparse in count and that will eventually be a set back for the deep learning models. The previous scenario can be handled by augmenting the dataset. Unlike augmentation of the dataset for image classification problem, this dataset also contains the annotated json files for each image. Hence we initially created mask RLE images from the annotated json file, then augmented each image along with its masks. We use Augmentor framework for augmenting both images and masks using two different operations, one is rotating the image and mask between 50 degree left and 50 degree right and another is flipping the images vertically. Using the two different operations we created 1200 augmented images from 410 original images

\section{Model}
\subsection{RCNN}
Unlike prior detection algorithms like overfeat \cite{overfeat} which uses sliding window protocol architecture for localizing the object, RCNN (Region Convolutional Neural Network) \cite{rcnn} uses combination of Region Proposal and CNN based architecture. The Region proposal head is used to select set of ROIs(Region of Interests) which has high possibility of having object in it form the image using category independent regional proposal algorithms like Objectness \cite{objectness}, Selective Search \cite{selectivesearch},..etc. Further n-dimensional feature map is been extracted from each ROI using set of convolutional layers (n varies with respect to the CNN architecture) which is followed by set of fully connected layers for classifying the ROI into its respected class. Even through this architecture seems to be simple, it won’t scale as there will be several thousand proposed ROIs from the Regional Proposed head and for all the ROIs features should be extracted using the CNN. This limitation is been overcome by other upcoming models which uses the outline of RCNN \cite{overfeat} architecture with modifications in Region proposed and CNN architecture.

\subsection{Faster RCNN}
RCNN \cite{rcnn} and Fast-RCNN \cite{frcnn}  are not fully differential because they use selective search for region proposals. RCNN performs selective search on the input image whereas Fast-RCNN \cite{frcnn} on convolutional features. Since the feature are richer in information and smaller in size, that helps Fast-RCNN to perform region proposals, an order of magnitude faster.

Faster-RCNN \cite{frrcnn} on the hand, use a learn-able Region proposal network which performs even more better. Region proposal network is a set of convolutional filters that detect the presence of an object(binary class score) and bounding box(x, y, h, w). Class score is casted as a classification problem and Bounding box is casted as a regression problem.

Since RPN can propose large number of regions for detection most of them overlap by large margin, Non-max suppresion (NMS) is used to reduce the number of proposals. The IoU, intersection over union score is used to determine how closer two proposals are and if the IOU score is less than 0.5, the regions are considered distinct and when the IoU is greater than 0.5 then the proposals are either merged or discarded based on the score.

\begin{figure}[h]
\includegraphics[width=85mm, height=40mm]{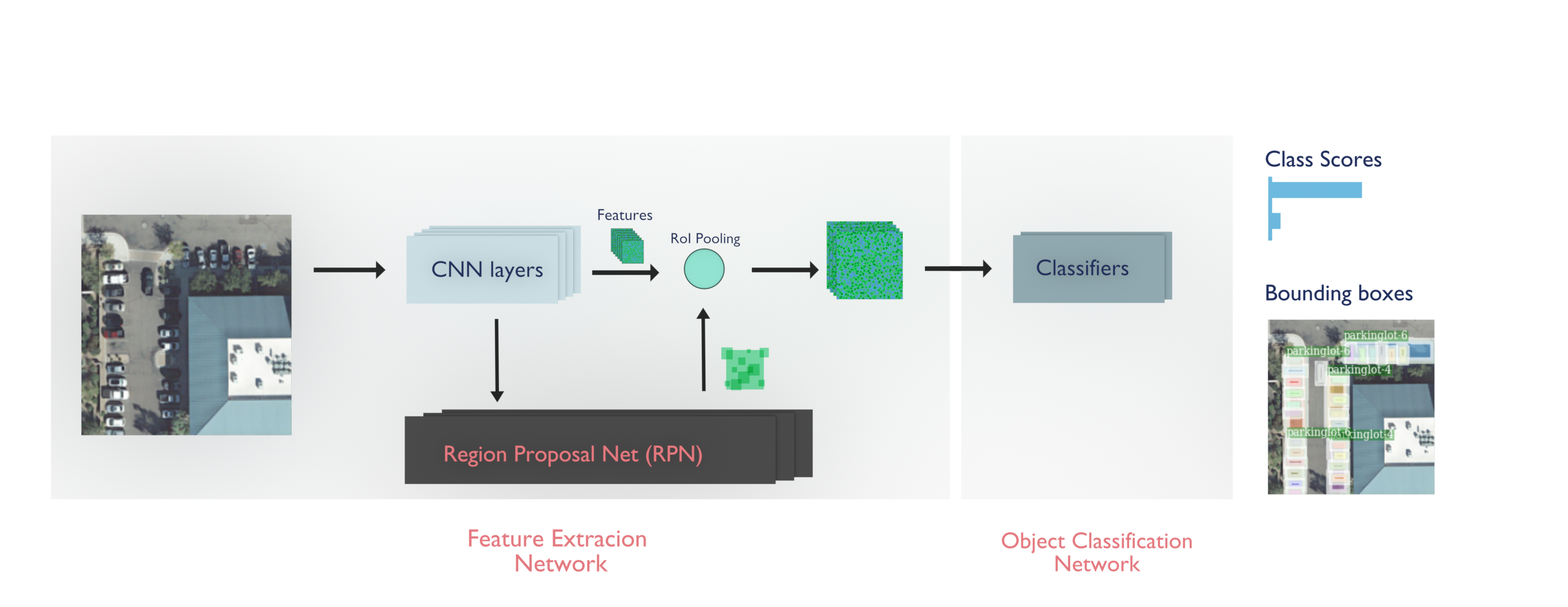}
\caption{Faster-RCNN}
\label{fig:model}
\end{figure}

\subsection{Mask RCNN}
Mask-RCNN \cite{mrcnn} extends Faster-RCNN \cite{frrcnn} by adding a branch for predicting an object mask in parallel with the existing branch for bounding box recognition and class classification. This architecture is easy to train and add only a small overhead to the Faster-RCNN architecture. Faster-RCNN was not designed for pixel to pixel alignment between network inputs and outputs and this is most evident in how ROIPool in Faster-RCNN works that performs coarse spatial quantization for feature extraction. To fix this mix alignment, Mask-RCNN \cite{mrcnn} uses a simple, quantization free layer called ROIAlign that preserves the exact spatial locations. ROIAlign improves the mask accuracy from 10\% to 50\%. Next vital change is decoupling mask generation with class prediction.

\begin{figure}[h]
\includegraphics[width=85mm, height=45mm]{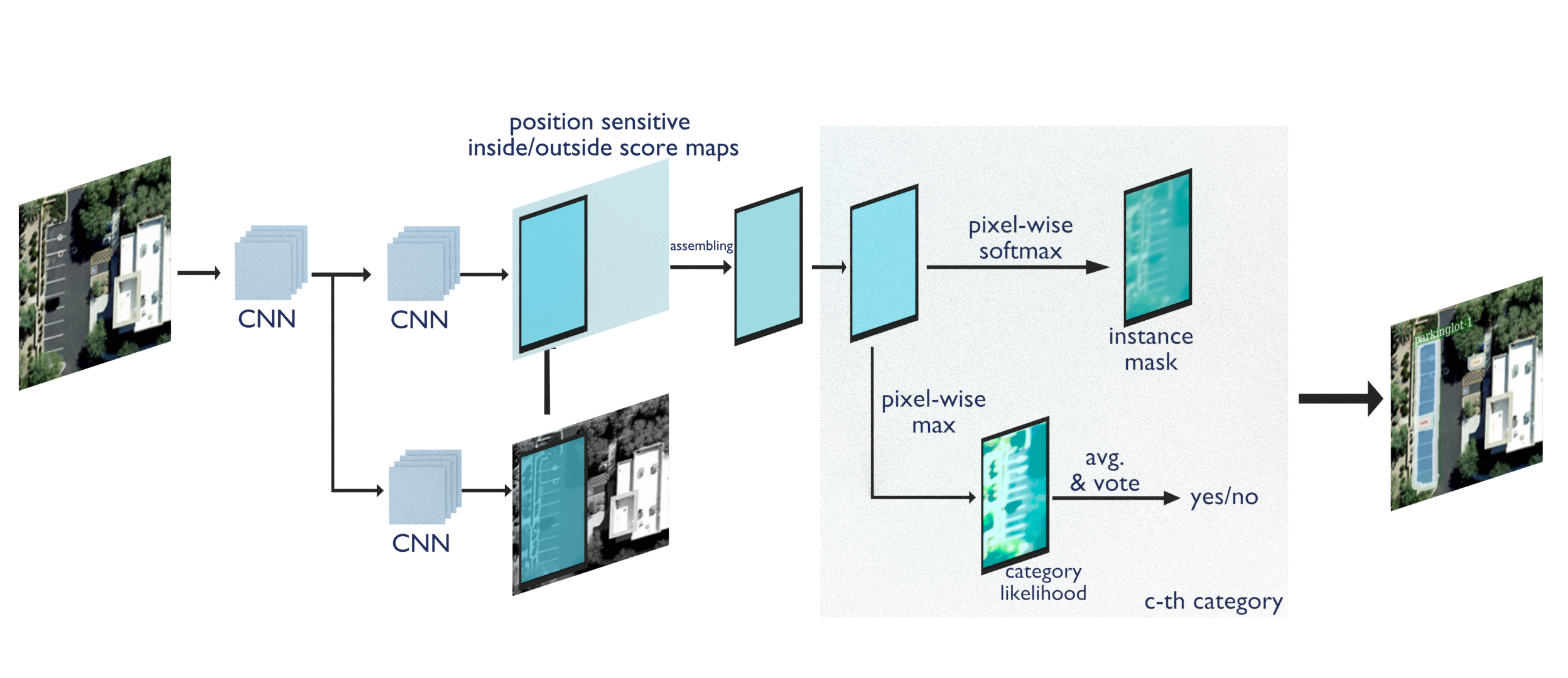}
\caption{Mask-RCNN}
\label{fig:model}
\end{figure}

\subsection{FPN}
Unlike a single feature map from the last convolutional layer of the model, a feature pyramid is build from the outputs of multiple convolution blocks. The feature pyramids are then used for object recognition as a way to handle scale invariance \cite{scaleinvar}. But recent object detection models have avoided pyramid representation as they are both computationally expensive and memory intensive, leaving the model to suffer from the scale invariance \cite{scaleinvar}. This scenario is been handled by the Feature Pyramid Networks(FPN) \cite{fpn}. Feature pyramids are collections of features computed at multi-scale versions of the same image. Improving on a similar idea proposed in DeepMask \cite{deepmask}, FPN brings backs feature pyramids using different feature maps of conv layers with differeent spatial resolutions with prediction happening on all levels of the pyramid. Using feature maps directly as it is would be tough as initial layers tend to contain lower level representations and poor semantics but good localisation whereas deeper layers tend to constitute higher level representations with rich semantics but suffer poor localization due to multiple subsampling.

\begin{figure}[h]
\includegraphics[width=80mm, height=80mm]{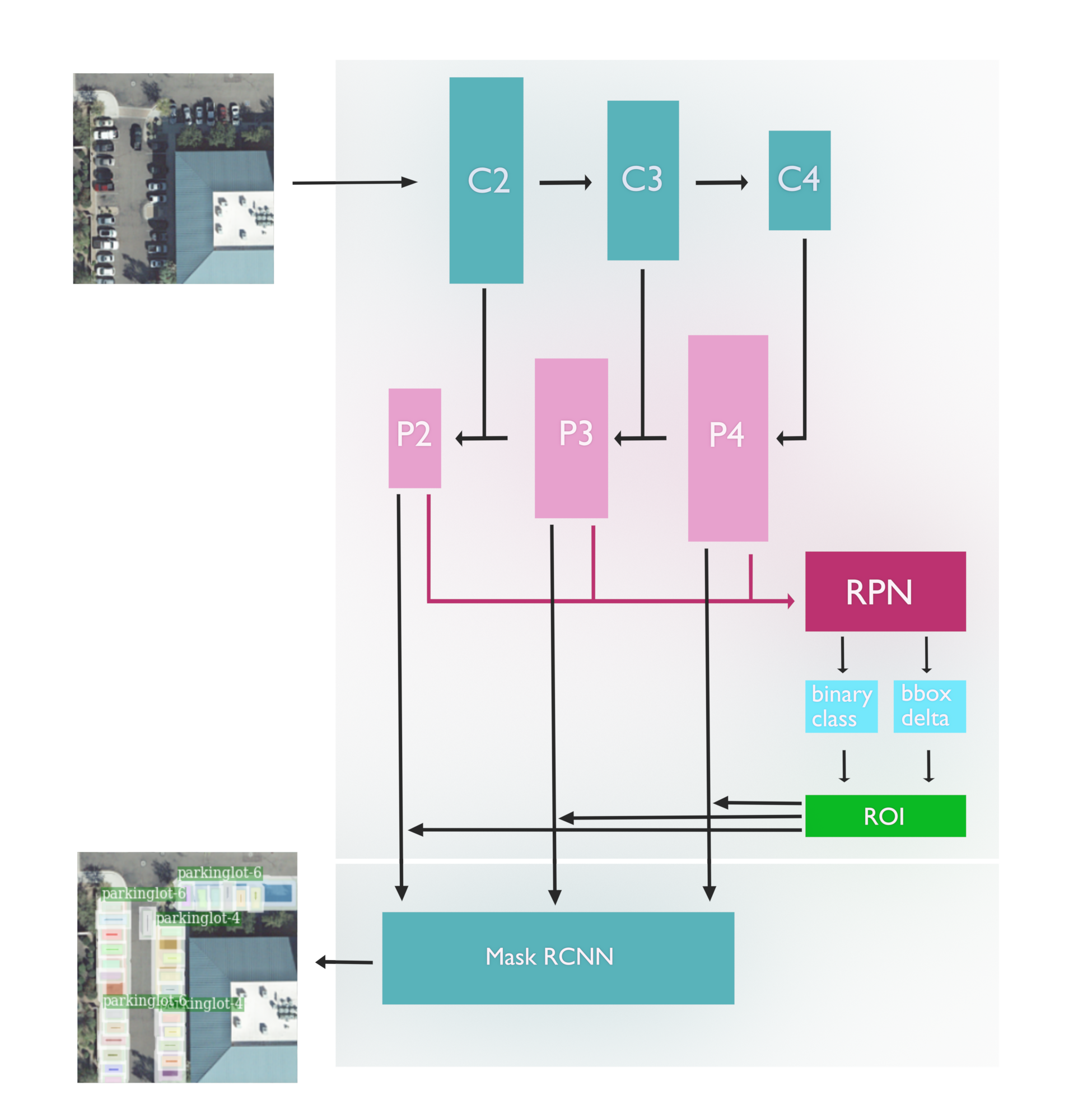}
\caption{FPN based Mask-RCNN}
\label{fig:model}
\end{figure}

\begin{table*}[h]
\caption{Comparison of accuracy between different architectures}
\centering
\begin{tabular}{ | l | l | l | l | l | l | l | l | }
\hline
Model & Attribute & AP & AP50 & AP75 & APs & APm & API\\
\hline

Faster RCNN Resnet-50 & bbox & 0.5658 & 0.7550 & 0.6697 & 0.5015 & 0.6180 & 0.4239 \\    \hline
Mask RCNN -Resnet-101 & bbox & 0.5826 & 0.7585 & 0.7052 & 0.5325 & 0.6241 & 0.2911 \\    
 & mask & 0.5332 & 0.7515 & 0.6444 & 0.4671 & 0.6090 & 0.4663 \\ \hline
Mask RCNN-Resnet-101 + FPN & bbox & 0.5919 & 0.7586 & 0.7092 & 0.5336 & 0.6410 & 0.4000 \\
& mask & 0.5530 & 0.7548 & 0.6701 & 0.4872 & 0.6224 & 0.4663 \\ \hline

\end{tabular}
\label{tab:models}
\end{table*}
\section{Model}

\subsection{Experiment}
Initially Faster-RCNN was implemented to localize the car and the parking lot The model takes the input image of  size W x H (width x Height) then computes the feature map using ResNet \cite{resnet} 50 Model (50-layer Residual Network). Consecutively, we use the region proposal network (RPN), which generates k anchors for each location. In our work, we took 3 different scales (128, 256, 512) with 3 different aspect ratio (1:2, 1:1, 2:1) which gives 9 anchors for each location. The RPN returns 2 sets of outputs, one is the objectness probability from the softmax layer (2k) and another is the bounding box coordinates (center-x, centre-y, width, height)  from the regressor layer (4k).
Thereafter, we sort the outputs from the RPN based on the objectness() and select first 15k values. Non Maximum Suppression (NMS) is then used on multiple candidates till all the duplicates for a single objects have been removed (eg: for a particular scenario, the 15k proposal anchors are reduced to 500 using NMS). 
Furthermore, each value in the remaining bbox coordinates and its respective region is cropped in the feature map . It is then resized to a uniform size of (32 x 32) with the use of RoI pooling. Then the resized feature map is then passed through a classification network with a 2 blocks of convolution networks and pooling layer. Lastly a linear layer is implemented for classifying the object class.
\\
The occupancy value deduced from the intersection of the vehicle to the parking area is not accurate. This is because the bounding box will not provide the exact area of the object. In order to account for the previous scenario we segmented both car and the parking lot by pixelwise using Mask-RCNN model which is a similar to the Faster-RCNN modeL. The only difference exhibited is that it has separate mask generation branch in parallel to the classification and bounding box detection.Therefore the usage of the mask for each and every instance of the object, the occupancy for each and every parking lot in a parking space can be determined. 
As there are spatial in-variance between objects in the parcel raster, FPN based Mask-RCNN would be more optimal than the normal mask-RCNN. After adding FPN based RPN proposals, the average precision for object both small and large area had increased considerably.
\\
We trained the Faster RCNN model with 1080 images for 40K steps using Stochastic gradient descent with 2 images in a batch in a single GTX 1080 Ti GPU for nearly 14 hrs. Same number of images are used for training Mask RCNN model which ran for 55K steps with same batch size in same single GPU for 23 hrs. For training Mask RCNN-FPN with the same dataset we used 3GTX  1080 Ti GPU machine with batch size of 6 for 10 hrs.

\section {Results and Analysis}

\begin{figure}[h]
\centering
\includegraphics[width=85mm, height=50mm]{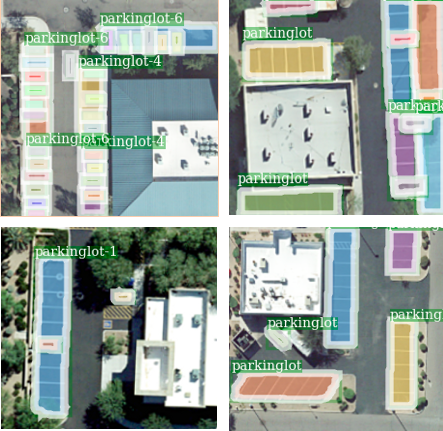}
\caption{Results of identification of parking spaces}
\label{fig:model}
\end{figure}

We ran three different experiments, one which only localize the parking space and vehicle with bounding box (bbox). The others were localized to the parking space and the vehicle with both bounding box and also the mask for that particular instance of the object. Average precision for different criteria are used for evaluate either the bounding box or the mask given for the objects but the models. Average precision is the mean of the precision score after detecting each single object in a image sorted according to the confidence value. The precision depends on the whether a predicted bbox or mask is True or False. For this we IOU(Intersection over union) metric and fix a threshold (eg: 0.5). So whenever a IOU of a bbox or mask is greater than the threshold the prediction is marked as True, the threshold may vary. By altering the criteria for both selecting the object and the threshold for IOU, we evaluated all the three models with 6 different metrics as a standard coco evaluation format. 

AP - average precision when ( 10 IOU is between 0.50 and 0.95 with difference of 0.05)
Ap50 - average precision when IOU is 0.50
AP75 - average precision when IOU is 0.75
APs - Average precision for small size object (area of the object is less than $32^\textsuperscript{2}$)
APm - Average precision for medium size object(area of the object between $32^\textsuperscript{2}$ and $96^\textsuperscript{2}$)
APl = Average precision for large size object (area of the object above $96^\textsuperscript{2}$)

From the results, its clear that the bbox detected by Faster RCNN is more accurate for large object compared to Mask RCNN but gradually reduced for smaller ones. Considering the Mask Accuracy, the FPN based Mask RCNN outperforms conventional Mask-RCNN in all metrics even it is been trained for lesser steps compared to the all models. By enriching the dataset with more scale invariance images and the training steps, the accuracy of the FPN based Mask RCNN will also be increased.

\begin{figure}[h]
\includegraphics[width=80mm, height=40mm]{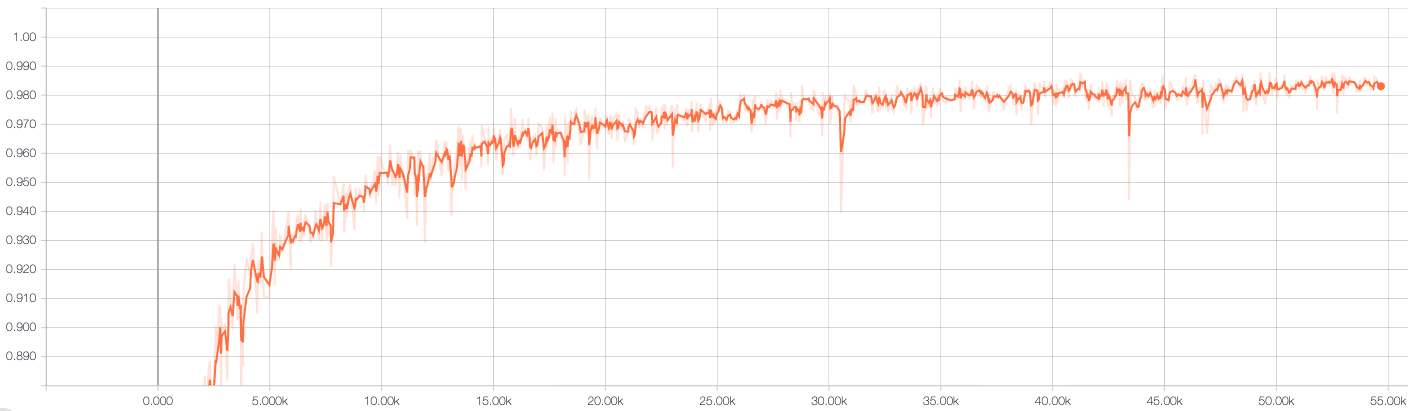}
\caption{Accuracy of Mask RCNN Model}
\label{fig:model}
\end{figure}

\begin{figure}[h]
\includegraphics[width=80mm, height=40mm]{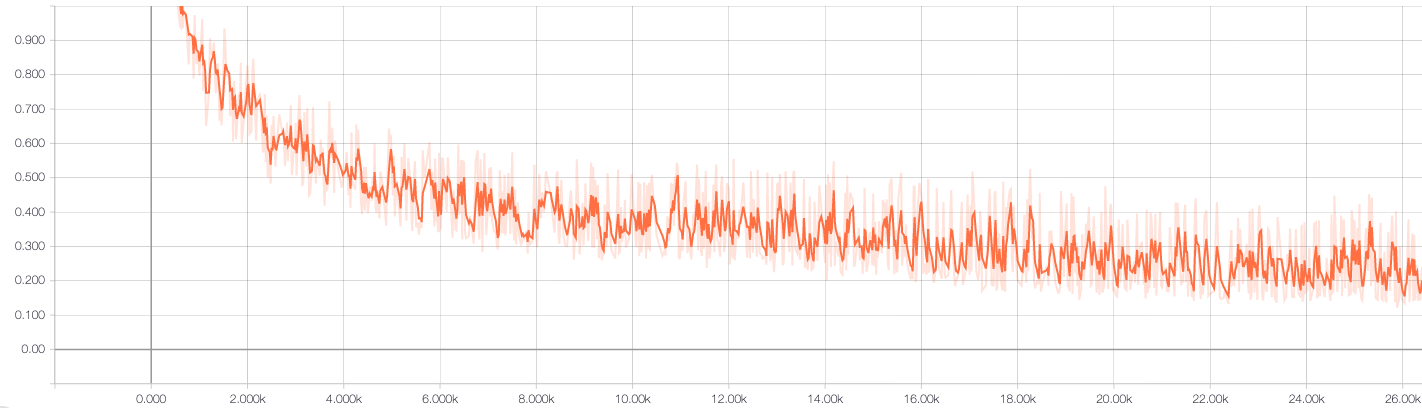}
\caption{Loss of Mask-RCNN Model}
\label{fig:model}
\end{figure}

\section{Conclusion}
From the results, its clear that the bbox detected by Faster RCNN is more accurate for large object compared to Mask RCNN but gradually reduced for smaller ones. Considering the Mask Accuracy, the FPN based Mask RCNN outperforms conventional Mask-RCNN in all metrics even it is been trained for lesser steps compared to the all models. Further increasing the dataset with more scale invariance images and the training steps, the accuracy of the FPN based Mask RCNN will also be increased. 

\section*{Acknowledgments}
The authors would like to thank Bhuvana Kundumani for reviewing the manuscript and for providing her technical inputs. The authors would also like to extend their gratitude to Saama Technologies Inc. for providing the perfect research and innovation environment.
\bibliographystyle{IEEEtran}

\begin{thebibliography}{29}

\bibitem{ultrasensor}
Pohl, J., Sethsson, M., Degerman, P.; Larsson, J. A semi-automated parallel parking system for passenger cars. J. Autom. Eng. 2006, 220, 53–65.

\bibitem{lasersensor}
Jung, H.G., Cho, Y.H., Yoon, P.J.; Kim, J. Scanning laser radar-based target position designation for parking
aid system. IEEE Trans. Intell. Transp. Syst. 2008, 9, 406–424.

\bibitem{cite1}
Real-time image-based parking occupancy detection using deep learning 
Debaditya Acharya Weilin Yan
Infrastructure Engineering, The University of Melbourn.

\bibitem{sift}
David Lowe, G. Distinctive Image Features from Scale-Invariant Keypoints, 2004.

\bibitem{pca}
Jollife, I., T. Principal Component Analysis and Factor Analysis. In: Principal Component Analysis. Springer Series in Statistics (1986). Springer, New York, NY

\bibitem{cite2}
Parking Space Classification using Convoluional Neural Networks
Jordan Cazamias \& Martina Marek
Stanford University
jaycaz@stanford.edu \& mmarek@cs.stanford.edu

\bibitem{satsensor}
Roy ,P.S., Behera, M.D, Srivastav, S.K., Satellite Remote Sensing: Sensor, Application and Techniques (2017), Proceedings of the National Academy of Sciences, India Section A: Physical Sciences.

\bibitem{orthorec}
Boccardo, P., Borgogno Mondino, E., Giulio Tonolo, F,. Lingua, A,. Orthorectification of High Resolution Satellite Images.

\bibitem{voc}
Everingham, M., Eslami, S.M.A., Van Gool, L., Williams, C.K.I., Winn, J., Zisserman, A. The PASCAL visual Object Classes Challenge: A Retrospective.

\bibitem{coco}
Tsung-Yi Lin, Michael Maire, Serge Belongie, Lubomir Bourdev, Ross Girshick, James Hays, Pietro Perona, Deva Ramanan, Lawrence Zitnick, C., Piotr Dollar. Microsoft COCO: Common Object in Context.

\bibitem{gdal}
GDAL/OGR contributors (2018). GDAL/OGR Geospatial Data Abstraction software Library. Open Source Geospatial Foundation. http://gdal.org

\bibitem{qgis}
QGIS Development Team (YEAR). QGIS Geographic Information System. Open Source Geospatial Foundation Project. http://qgis.osgeo.org

\bibitem{labelme}
Russell, B., C., Torralba, A., Murphy, K., P., Freeman, W., T. LabelMe: a database and web-based tool for image annotation. International Journal of Computer Vision, pages 157-173, Volume 77, Numbers 1-3, May, 2008.

\bibitem{overfeat}
Pierre Sermanet, David Eigen, Xiang Zhang, Michael Mathieu, Rob Fergus, Yann LeCun. Overfeat: Integrated Recognition, Localization and Detection using Convolutional Features (2014).

\bibitem{rcnn}
Ross Girshick, Jeff Donahue, Trevor Darrell, Jitendra Malik. Rich Feature Hierarchies for accurate object detection and sematic segmentation (2014).

\bibitem{objectness}
Alexe, B., Deselaers, T., Ferrari, V,. Measuring the objectness of image windows. TPAMI, 2012.

\bibitem{selectivesearch}
Uijlings, G., Van De Sande, K., Gevers, T., Smeulders, A., Selective Search for object recognition. IJCV, 2013.

\bibitem{cnn}
Yann LeCun, Patrick Haffner, Leon Bottou, Yoshua Bengio. Object Recognition with Gradient Based Learning (1999).

\bibitem{frcnn}
Ross Girshick. Fast R-CNN (2015). 

\bibitem{frrcnn}
Shaoqing Ren, Kaiming He, Ross Girshick, Jian Sun. Faster R-CNN: Towards Real-Time Object Detection with Region Proposal Networks (2016).

\bibitem{mrcnn}
Kaiming He, Georgia Gkiozari, Piotr Dollar, Ross Girshick. Mask R-CNN (2018).

\bibitem{fpn}
Tsung-Yi Lin, Piotr Dollar, Ross Girshick, Kaiming He, Bharath Hariharan, Serge Belongie. Feature Pyramid Networks for Object Detection (2017).

\bibitem{deepmask}
Pedro Pinheiro, O., Ronan Collobert, Piotr Dollar. Learning to Segment Object Candidates, 2015. Facebook AI Research.

\bibitem{resnet}
Kaiming He, Xiangyu Zhang, Shaoqing Ren, Jian Sun. Deep Residual Learning for Image Recognition, 2015. Microsoft Research.

\bibitem{scaleinvar}
Bharat Singh, Larry Davis, s,. An Analysis of scale Invariance in Object Detection - SNIP(2018).
\
\end{thebibliography}

\end{document}